\documentclass[times, twocolumn, final]{elsarticle}

\usepackage{framed,multirow}
\usepackage{amssymb}
\usepackage{latexsym}
\usepackage{url}
\usepackage{xcolor}
\usepackage[cmex10]{amsmath}
\usepackage{bm}
\usepackage{subfigure}
\usepackage{soul}

\usepackage{lineno,hyperref}
\modulolinenumbers[5]

\journal{arXiv}

\biboptions{authoryear}

\begin{document}

\begin{frontmatter}

\title{Unsupervised shape and motion analysis of 3822 cardiac 4D MRIs \\of UK Biobank}

\author[1]{Qiao Zheng\corref{mycorrespondingauthor}}
\cortext[mycorrespondingauthor]{Corresponding author}
\ead{qiao.zheng@inria.fr}

\author[1]{Herv\'{e} Delingette}
\author[2,3]{Kenneth Fung}
\author[2,3]{Steffen E. Petersen}
\author[1]{Nicholas Ayache}

\address[1]{Universit\'{e} C\^{o}te d'Azur, Inria, 2004 Route des Lucioles, 06902 Sophia Antipolis, France}

\address[2]{William Harvey Research Institute, NIHR Barts Biomedical Research Centre, Queen Mary University of London, London EC1M 6BQ, UK}

\address[3]{Barts Heart Centre, St Bartholomew’s Hospital, Barts Health NHS Trust, London EC1A 7BE, UK}


\begin{abstract}
We perform unsupervised analysis of image-derived shape and motion features extracted from 3822 cardiac 4D MRIs of the UK Biobank. First, with a feature extraction method previously published based on deep learning models, we extract from each case 9 feature values characterizing both the cardiac shape and motion. Second, a feature selection is performed to remove highly correlated feature pairs. Third, clustering is carried out using a Gaussian mixture model on the selected features. After analysis, we identify two small clusters which probably correspond to two pathological categories. Further confirmation using a trained classification model and dimensionality reduction tools is carried out to support this discovery. Moreover, we examine the differences between the other large clusters and compare our measures with the ground-truth.

\end{abstract}


\end{frontmatter}


\section{Introduction}
\label{sec1}

In recent years, more and more data are made accessible for research in medical image analysis. For instance, the UK Biobank study of \cite{Petersen:2017} has released a dataset containing the cardiac cine MRI images of thousands of volunteers, from which various key cardiovascular functional indexes can be extracted for analysis (\cite{Attar:2019}). The Alzheimer's Diseases Neuroimaging Initiative (ADNI, \cite{Toga:2015}) has accumulated brain scan images of about two thousand participants. The abundant data available in the community is certainly a highly valuable resource (\cite{Rueckert:2016}, \cite{Suinesiaputra:2016}). Researchers are hence less constrained by the scarcity of data which has been a prevailing challenge for a long time. Further research is necessary (\cite{Zhang:2016}, \cite{Barillot:2016}) on new topics associated with big data. For example, one major challenge is how to make good use of unlabeled data (\cite{Bruijne:2016}, \cite{Weese:2016}). In fact, while there are more and more labeled data available, an important part of medical images are still unlabeled. This is understandable as it is in general expensive and tedious to diagnose and label cases by human experts. Methods that can extract useful information from unlabeled data are hence interesting and might potentially save a lot of time and effort.

Many research projects have been developed to perform pathology-related analysis using features extracted from medical images. Many of these works focus on brain scan images. For example, in \cite{Parisot:2018}, feature vectors extracted from brain images are used for the prediction of autism spectrum disorder and Alzheimer's disease. An anatomical landmark based deep feature representation for MRI is proposed in \cite{Liu:2018} for diagnosis of brain disease. Some other studies are based on digital histopathological images. For instance, \cite{Madabhushi:2017} discusses the predictive modeling of digital histopathological images from a detection, segmentation, feature extraction, and tissue classification perspective. \cite{Komura:2018} reviews the machine learning methods for histopathological image analysis. But there are less pathology-related and feature-based researches on cardiac images than on brain scan images and digital histopathological images. And currently, this research (\cite{Zheng:2018:2}, \cite{Khened:2018}, \cite{Khened:2017}, \cite{Isensee:2017}, \cite{Wolterink:2017}, \cite{Cetin:2017}) is mostly about pathology classification in the dataset of Automatic Cardiac Diagnosis Challenge (ACDC) of MICCAI 2017, which contains 100 cases with labels. The work of \cite{Attar:2019} is one of the very first projects to propose a fully automatic, high throughput image parsing workflow for the analysis of cardiac MRI in UK Biobank with systematic tests of the performance. As an extension of the previous works and a challenge to ourselves, we wish to conduct unsupervised analysis on large unlabeled cardiac image datasets.

Clustering, an unsupervised machine learning technique that groups similar entities together, might be suitable for analyzing large unlabeled datasets. Up to now, clustering has been widely used on image segmentation in medical image analysis. For example, the authors of \cite{Kinani:2017} develop a tool based on clustering to outline brain lesion contours. Unsupervised segmentation of 3D lung CT images is proposed in \cite{Moriya:2018} based on clustering and deep representation learning. Some studies show that clustering is also a powerful tool for classification. For instance, a clustering method is applied to classify the analyzed brain images into healthy and multiple sclerosis disease in \cite{Moldovanu:2015}. The authors of \cite{Kawadiwale:2014} introduce various clustering techniques to classify brain MR images into normal and malformed. While most of the application of clustering in the domain is on brain images, we aim to extend its application to cardiac images.

In this paper, we perform a cluster analysis of a group of features extracted from the cardiac MR images of the UK Biobank dataset. Our main contributions are threefold: \\
\textbullet \ We conduct a cardiac-pathology-related analysis on a large unlabeled dataset. \\
\textbullet \ As a novel application of a classic method in medical image analysis, clustering is used in our analysis to group cases without supervision. \\
\textbullet \ Among the resulting clusters, two can indeed be identified as leaning toward pathological categories.

\section{Data}

\subsection{UK Biobank}
The proposed method was applied to the very large UK Biobank cardiac MRI dataset, see  \cite{Petersen:2016}\footnote{Application Number 2964.}. It comprises short-axis cine MRI of about five thousand participants from the general population. More details of the magnetic resonance protocol are available in \cite{Petersen:2016}. Each time series consists of 3D volumes with slice thickness of 8mm for short-axis images and 6mm for long-axis images. The in-plane resolution is 1.8mm $\times$ 1.8mm. Volumes at end-diastole (ED) and end-systole (ES) and ejection fraction for left ventricle cavity (LVC) were derived from InlineVF analysis algorithm (\cite{Jolly:2013}, \cite{Lu:2010}) performed by UK Biobank (Field 22421-22422). Those values are considered in this paper as ground-truth (or reference) values. To be consistent with our previous research such as \cite{Zheng:2018} and \cite{Zheng:2018:2}, we exclude roughly one thousand cases that are provided with incomplete or unconvincing ground-truth. The remaining 3822 cases are then used for cluster analysis. For part of these cases, the measures of LVC volumes at ED and ES and LVC ejection fraction are provided as ground-truth by UK Biobank.

As pointed out on the website of UK Biobank\footnote{https://www.ukbiobank.ac.uk/scientists-3/} and in \cite{Fry:2017}, while UK Biobank participants are not representative of the general population with evidence of a `healthy volunteer' selection bias (and hence cannot be used to provide representative disease prevalence and incidence rates), valid assessment of exposure-disease relationships are nonetheless widely generalizable and does not require participants to be representative of the population at large. 

\subsection{ACDC}
In the experiment part, we will show the correspondence between some resulting clusters and the definition of some pathology categories defined in the ACDC challenge. Furthermore, a classification model trained on ACDC by \cite{Zheng:2018:2} will be applied on UK Biobank for comparison with the clustering method proposed in this paper. The ACDC challenge dataset consists of 100 cases, which are divided into 5 pathological groups of equal size according to their pathology on either the left ventricle (LV) or the right ventricle (RV): \\
\textbullet \ dilated cardiomyopathy (DCM): left ventricle cavity (LVC) volume at ED larger than 100 $\mathit{mL/m^2}$ and LVC ejection fraction lower than 40\% \\ 
\textbullet \ hypertrophic cardiomyopathy (HCM): left ventricle (LV) cardiac mass higher than 110 $\mathit{g/m^2}$, several myocardial segments with a thickness higher than 15 mm at ED and a normal ejection fraction \\ 
\textbullet \ myocardial infarction (MINF): LVC ejection fraction lower than 40\% and several myocardial segments with abnormal contraction\\
\textbullet \ RV abnormality (RVA): right ventricle cavity (RVC) volume higher than 110 $\mathit{mL/m^2}$ or RVC ejection fraction lower than 40\% \\
\textbullet \ normal subjects (NOR) \\

\section{Methods}

There are mainly three steps in the proposed method: feature extraction, feature selection and cluster analysis.

\subsection{Feature Extraction}

\begin{table}[]
\caption{The 9 features generated by our feature extraction method. Among them 8 are selected for cluster analysis.}
\centering
\begin{tabular}{c c c}
\hline
\noalign{\vskip 0.0in}
\multicolumn{1}{|c}{Feature} & \multicolumn{1}{|c|}{Notion} & \multicolumn{1}{c|}{Selected} \\ 
\hline
\multicolumn{1}{|c}{RVC volume at ED} & \multicolumn{1}{|c|}{$V_\mathit{RVC,ED}$} & \multicolumn{1}{c|}{yes}\\ 
\hline
\multicolumn{1}{|c}{LVC volume at ES} & \multicolumn{1}{|c|}{$V_\mathit{LVC,ES}$} & \multicolumn{1}{c|}{yes}\\ 
\hline
\multicolumn{1}{|c}{RVC ejection fraction} & \multicolumn{1}{|c|}{$\mathit{EF_{RVC}}$ } & \multicolumn{1}{c|}{yes}\\ 
\hline
\multicolumn{1}{|c}{LVC ejection fraction} & \multicolumn{1}{|c|}{$\mathit{EF_{LVC}}$ } & \multicolumn{1}{c|}{no}\\ 
\hline
\multicolumn{1}{|c}{Ratio between RVC and} & \multicolumn{1}{|c|}{$\mathit{R_{RVCLV,ED}}$} & \multicolumn{1}{c|}{yes}\\ 
\multicolumn{1}{|c}{LV volumes at ED} & \multicolumn{1}{|c|}{} & \multicolumn{1}{c|}{}\\ 
\hline
\multicolumn{1}{|c}{Ratio between LVM and} & \multicolumn{1}{|c|}{$\mathit{R_{LVMLVC,ED}}$} & \multicolumn{1}{c|}{yes}\\ 
\multicolumn{1}{|c}{LVC volumes at ED} & \multicolumn{1}{|c|}{} & \multicolumn{1}{c|}{}\\ 
\hline
\multicolumn{1}{|c}{Maximal LVM thickness} & \multicolumn{1}{|c|}{$\mathit{MT_{LVM,ED}}$} & \multicolumn{1}{c|}{yes} \\ 
\multicolumn{1}{|c}{in all the slices at ED} & \multicolumn{1}{|c|}{} & \multicolumn{1}{c|}{}\\ 
\hline
\multicolumn{1}{|c}{Radius motion} & \multicolumn{1}{|c|}{$\mathit{RMD}$} & \multicolumn{1}{c|}{yes}\\ 
\multicolumn{1}{|c}{disparity} & \multicolumn{1}{|c|}{} & \multicolumn{1}{c|}{}\\ 
\hline
\multicolumn{1}{|c}{Thickness motion} & \multicolumn{1}{|c|}{$\mathit{TMD}$} & \multicolumn{1}{c|}{yes}\\ 
\multicolumn{1}{|c}{disparity} & \multicolumn{1}{|c|}{} & \multicolumn{1}{c|}{}\\ 
\hline
\end{tabular}
\end{table}

The feature extraction method used in this paper is the same as the one proposed in our previous work published by \cite{Zheng:2018:2}. We briefly describe its principal steps again below.

The first part of the feature extraction method generates 7 shape-related features. Segmentation with spatial propagation has been proven to be consistent and robust (\cite{Zheng:2018}, \cite{Zheng:2018:3}). With the cardiac segmentation method proposed in \cite{Zheng:2018}, the cardiac images are segmented such that we obtain the masks of LVC, left ventricle myocardium (LVM) and RVC on both ED and ES frames. Then the volumes of LVC, LVM and RVC at both ED and ES can be computed directly, as can the thickness of LVM. Finally, 7 shape-related features are generated (the first 7 terms in Table 1).

The second part of the method extracts 2 motion-characteristic features. Using a neural network which outputs apparent flow maps given image pairs, we get a series of apparent flow maps characterizing the in-plane motion for each MRI slice of each case. Combined with the LVM segmentation mask obtained as described above, the motion of each myocardium pixel is hence available. Eventually, 2 features are computed to present the disparity of the radial myocardial motion and the myocardial thickening respectively (the last 2 rows in Table 1).

In total, from the images of each case, 9 features characterizing the shape and the motion of the heart are extracted.

\subsection{Feature Selection}

As shown in \cite{Zheng:2018:2}, these extracted features can be used for cardiac pathology classification in the ACDC dataset with performances comparable to the state-of-the-art. However, these features are not necessarily independent. Some might be redundant if there are highly correlated feature pairs. In cluster analysis, if too many variables are used simultaneously, the redundant ones serve only to create noise that harms the clustering. So it is helpful to select a sub-group of features by removing highly correlated feature pairs.

For each pair among the 9 extracted features, we compute the Pearson correlation coefficient and the maximal information coefficient (MIC) (\cite{Reshed:2011}). The former measures the linear correlation between two features, while the latter measures the mutual information between features. If there is any highly correlated pair according to these measures (i.e. Pearson correlation coefficient of absolute value above 0.8, or MIC above 0.5), we will exclude one feature in this pair. The remaining features are then considered as selected.

\subsection{Cluster Analysis}
We perform a model selection of Gaussian mixture model using the Bayesian information criterion (BIC). Then the selected Gaussian mixture model is applied to cluster the 8 selected features.

\subsubsection{Gaussian Mixture Model Selection}

A Gaussian mixture model (\cite{Reynolds:2009}) is a probabilistic model which assumes that the data points are generated from a mixture of a certain number of Gaussian distributions with unknown parameters. An expectation-maximization algorithm is used to iteratively estimate its parameters from data. Then the fitted model can assign to each sample the Gaussian component it most likely belongs to.

We use the Gaussian mixture model as implemented in scikit-learn (\cite{Pedregosa:2011}). It has two major parameters, the type of covariance matrix and the number of components, upon which a selection is necessary. For this purpose, we calculate the Bayesian information criterion (BIC, \cite{Wit:2012}) for Gaussian mixture models with different types of covariance matrix and numbers of components. In theory, BIC recovers the true number of components approximately. We fit the Gaussian mixture models with the following types of covariance matrix: \\
\textbullet \ `tied': all components share the same covariance matrix; \\
\textbullet \ `diag': each component has its own diagonal covariance matrix; \\
\textbullet \ `full': each component has its own covariance matrix. \\
The number of components is also varied. By looking for models with the smallest BIC scores, we wish to select the most simple model that can fit the data thereby idenitifying the most suitable type of covariance matrix and a range of reasonable numbers of components.

The number of components will finally be determined by examining the sizes of resulting clusters of the Gaussian mixture models. More details will be provided in the Experiments and Results section.

\subsubsection{Analysis of the Resulting Clusters}
The clusters generated by the selected model will be examined. In particular, we verify if the cases in any of the clusters correspond to a pathological category according to the definitions of pathologies given by the ACDC challenge.

\section{Experiments and Results}

\subsection{Feature Extraction}
With the feature extraction method introduced in the Methods section, for each of the 3822 UK Biobank cases, 9 feature values are extracted.

\subsection{Feature Selection}

\begin{figure}[t]
\centering
\includegraphics[width=8.7cm, height=7.0cm]{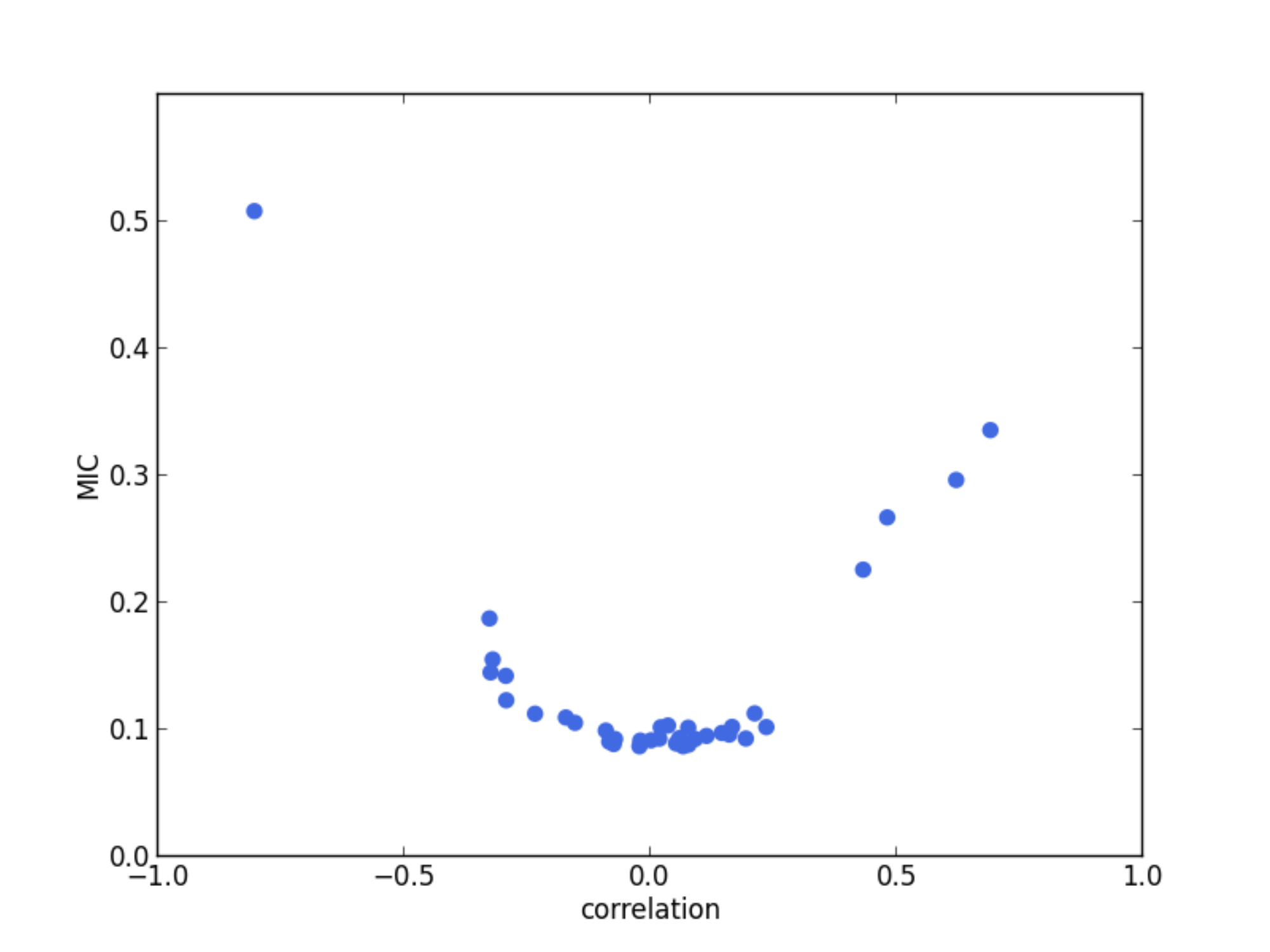}
\caption{Pearson correlation coefficient versus MIC. Each point corresponds to a pair of features. The point in the upper-left corner corresponds to $V_\mathit{LVC,ES}$ and $\mathit{EF_{LVC}}$. The strong negative correlation between these two features is reasonable, since by definition $\mathit{EF_{LVC}} = 1 - V_\mathit{LVC,ES}/V_\mathit{LVC,ED}$, in which $V_\mathit{LVC,ED}$ is the LVC volume at ED.}
\end{figure}

We calculate the Pearson correlation coefficient and MIC for each pair of features among the 9 extracted features. In Figure 1, the plot of Pearson correlation coefficient versus MIC, it is clear that the absolute value of the Pearson correlation coefficient and MIC are positively correlated. There is only one point on the upper left corner of the plot representing a highly correlated pair. It corresponds to $V_\mathit{LVC,ES}$ and $\mathit{EF_{LVC}}$, which are of Pearson correlation coefficient -0.80 and MIC 0.51. The strong negative correlation between these two features is reasonable, since by definition $\mathit{EF_{LVC}} = 1 - V_\mathit{LVC,ES}/V_\mathit{LVC,ED}$, in which $V_\mathit{LVC,ED}$ is the LVC volume at ED. Therefore, $V_\mathit{LVC,ES}$ and $\mathit{EF_{LVC}}$ appear to be redundant. Hence we exclude $\mathit{EF_{LVC}}$ and select the remaining 8 features for cluster analysis (Table 1).

\subsection{Cluster Analysis}

\subsubsection{Gaussian Mixture Model Selection}

\begin{figure}[t]
\centering
\includegraphics[width=8.7cm, height=7.0cm]{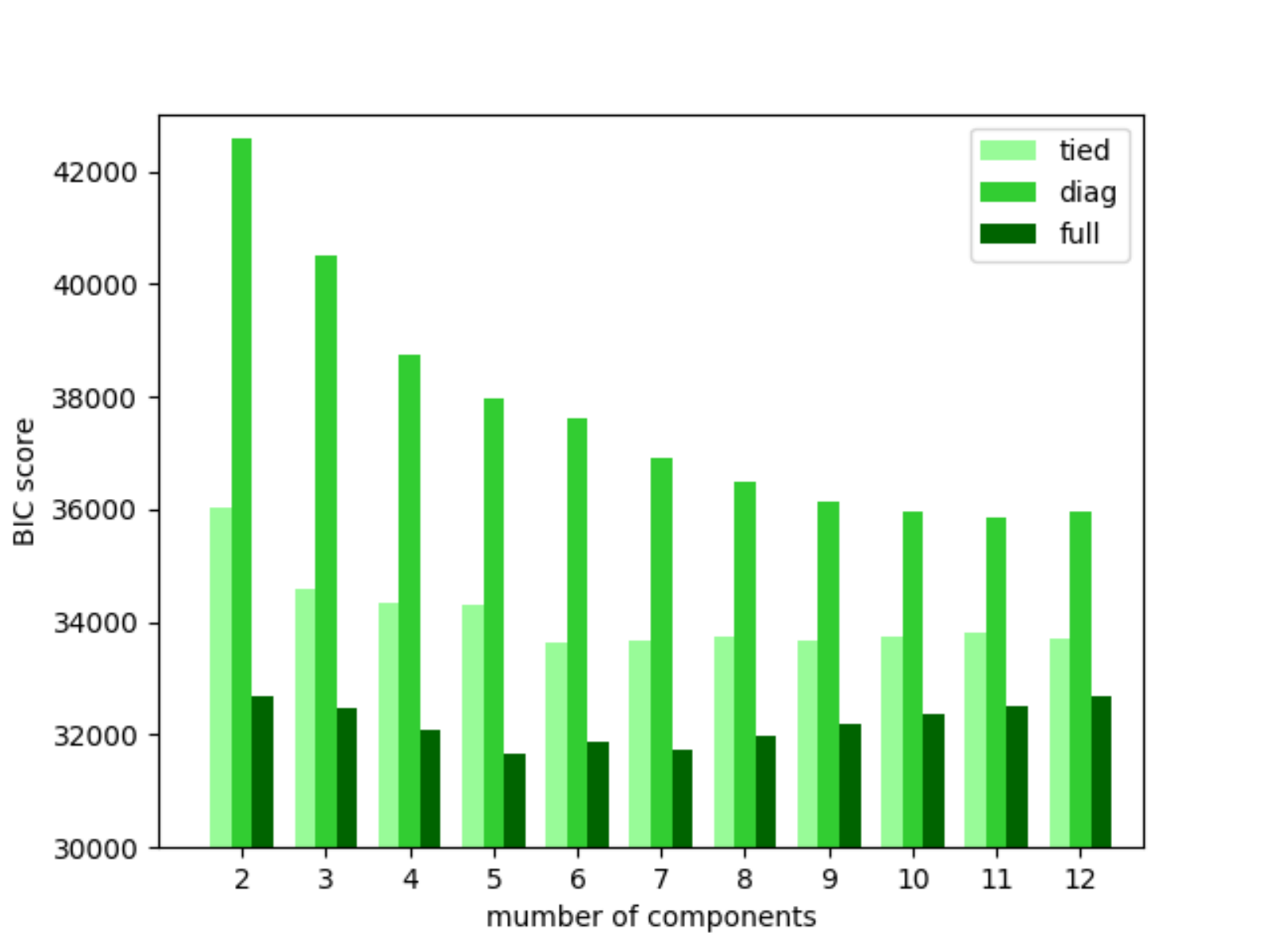}
\caption{BIC scores of Gaussian mixture models with various types of covariance matrix and numbers of components}
\end{figure}

The BIC scores of the Gaussian mixture models with various types of covariance matrix and numbers of components are plotted in Figure 2. It is clear that the `full' covariance matrix type is the best among the three. The `full' covariance matrix type is hence selected. 

And in terms of the number of components, the Gaussian mixture models with the `full' covariance matrix type of 3 to 10 components have the smallest BIC scores. Among them, we find that: \\
\textbullet \ The models of 3 to 6 components only generate large clusters, each of which contains at least about one hundred cases. \\
\textbullet \ The models of 7 and 8 components bring about only one small cluster (less than a dozen cases). \\
\textbullet \ The models of 9 and 10 components give rise to two small clusters (less than a dozen cases). \\

According to the statistics\footnote{https://www.bhf.org.uk/what-we-do/our-research/heart-statistics (accessed January 20, 2019)} provided by the British Heart Foundation , about 7 million people in the UK are living with cardiovascular diseases, which is about 10.6\% of the total population. More specifically, if we look at the most common cardiovascular disease categories, the percentages of UK population living with myocardial infarction, atrial fibrillation and heart failure are about 1.5\%, 2.0\% and 1.4\%, respectively. This means that most of the cases in the general population do not have a cardiac pathology. Taking the `healthy volunteer' selection bias of UK Biobank mentioned in Section 2.1 into account, the cases of cardiovascular diseases are hence probably exceedingly rare in UK Biobank. Thus, if there is any cluster that is related to a specific pathological category in an interpretable manner, its size should be small, say, no more than 76 (2\% of the 3822 UK Biobank cases).

So we can now suggest that a component number of 9 or 10 is probably most suitable. We choose the model of 9 components for further analysis. But we would like to point out that the two resulting small clusters of the models of 9 and 10 components are very similar in terms of size and cases. So the results and the conclusions shown below will be roughly the same if we use the model of 10 components.

To summarize, the Gaussian mixture model with the `full' covariance matrix type and 9 components is selected.

\subsubsection{Analysis of the Resulting Clusters}

\begin{table}[]
\caption{RVC volumes and ejection fraction at ED of the cases of cluster \#5 based on our feature extraction method.}
\centering
\begin{tabular}{c c c}
\hline
\noalign{\vskip 0.0in}
\multicolumn{1}{|c}{ID} & \multicolumn{1}{|c|}{RVC volume at ED} & \multicolumn{1}{c|}{RVC ejection}\\ 
\multicolumn{1}{|c}{} & \multicolumn{1}{|c|}{($\mathit{mL/m^2}$)} & \multicolumn{1}{c|}{fraction}\\ 
\hline
\multicolumn{1}{|c}{2512949} & \multicolumn{1}{|c|}{133.13} & \multicolumn{1}{c|}{63.61\%}\\ 
\hline
\multicolumn{1}{|c}{2628396} & \multicolumn{1}{|c|}{175.77} & \multicolumn{1}{c|}{43.91\%}\\ 
\hline
\multicolumn{1}{|c}{3423847} & \multicolumn{1}{|c|}{140.50} & \multicolumn{1}{c|}{65.24\%}\\ 
\hline
\multicolumn{1}{|c}{3713328} & \multicolumn{1}{|c|}{169.65} & \multicolumn{1}{c|}{71.59\%}\\ 
\hline
\multicolumn{1}{|c}{3874816} & \multicolumn{1}{|c|}{183.96} & \multicolumn{1}{c|}{56.22\%}\\ 
\hline
\multicolumn{1}{|c}{4366978} & \multicolumn{1}{|c|}{134.68} & \multicolumn{1}{c|}{52.53\%}\\ 
\hline
\multicolumn{1}{|c}{4681487} & \multicolumn{1}{|c|}{139.82} & \multicolumn{1}{c|}{54.39\%}\\ 
\hline
\multicolumn{1}{|c}{4710306} & \multicolumn{1}{|c|}{144.86} & \multicolumn{1}{c|}{29.69\%}\\ 
\hline
\multicolumn{1}{|c}{5101726} & \multicolumn{1}{|c|}{145.93} & \multicolumn{1}{c|}{43.82\%}\\ 
\hline
\multicolumn{1}{|c}{5319688} & \multicolumn{1}{|c|}{151.30} & \multicolumn{1}{c|}{51.93\%}\\ 
\hline
\multicolumn{1}{|c}{5561149} & \multicolumn{1}{|c|}{180.48} & \multicolumn{1}{c|}{41.88\%}\\ 
\hline
\end{tabular}
\end{table}

\begin{table*}[t]
\caption{LVC volumes at ED and ejection fraction of the cases of cluster \#8 based on our feature extraction method (the 2nd and 3rd columns). The same measures provided by the UK Biobank dataset are also shown (the 4th and 5th columns). The two sets of measures are quite close to each other.}
\centering
\begin{tabular}{c c c c c}
\hline
\noalign{\vskip 0.0in}
\multicolumn{1}{|c}{} & \multicolumn{1}{|c|}{} & \multicolumn{1}{c|}{} & \multicolumn{1}{c|}{Ground-truth} & \multicolumn{1}{c|}{Ground-truth}\\ 
\multicolumn{1}{|c}{ID} & \multicolumn{1}{|c|}{LVC volume at ED} & \multicolumn{1}{c|}{LVC ejection} & \multicolumn{1}{c|}{LVC volume at ED} & \multicolumn{1}{c|}{LVC ejection}\\
\multicolumn{1}{|c}{} & \multicolumn{1}{|c|}{($\mathit{mL/m^2}$)} & \multicolumn{1}{c|}{fraction} & \multicolumn{1}{c|}{($\mathit{mL/m^2}$)} & \multicolumn{1}{c|}{fraction}\\ 
\hline
\multicolumn{1}{|c}{2432774} & \multicolumn{1}{|c|}{189.28} & \multicolumn{1}{c|}{19.74\%} & \multicolumn{1}{c|}{208.24} & \multicolumn{1}{c|}{20\%} \\ 
\hline
\multicolumn{1}{|c}{3378112} & \multicolumn{1}{|c|}{213.28} & \multicolumn{1}{c|}{18.75\%} & \multicolumn{1}{c|}{213.03} & \multicolumn{1}{c|}{15\%} \\ 
\hline
\multicolumn{1}{|c}{4879002} & \multicolumn{1}{|c|}{133.09} & \multicolumn{1}{c|}{27.03\%} & \multicolumn{1}{c|}{144.59} & \multicolumn{1}{c|}{29\%} \\ 
\hline
\multicolumn{1}{|c}{5618713} & \multicolumn{1}{|c|}{192.87} & \multicolumn{1}{c|}{26.74\%} & \multicolumn{1}{c|}{192.43} & \multicolumn{1}{c|}{27\%} \\ 
\hline
\end{tabular}
\end{table*}

\begin{figure*}[]
\centering
\includegraphics[width=16.1cm, height=8.3cm]{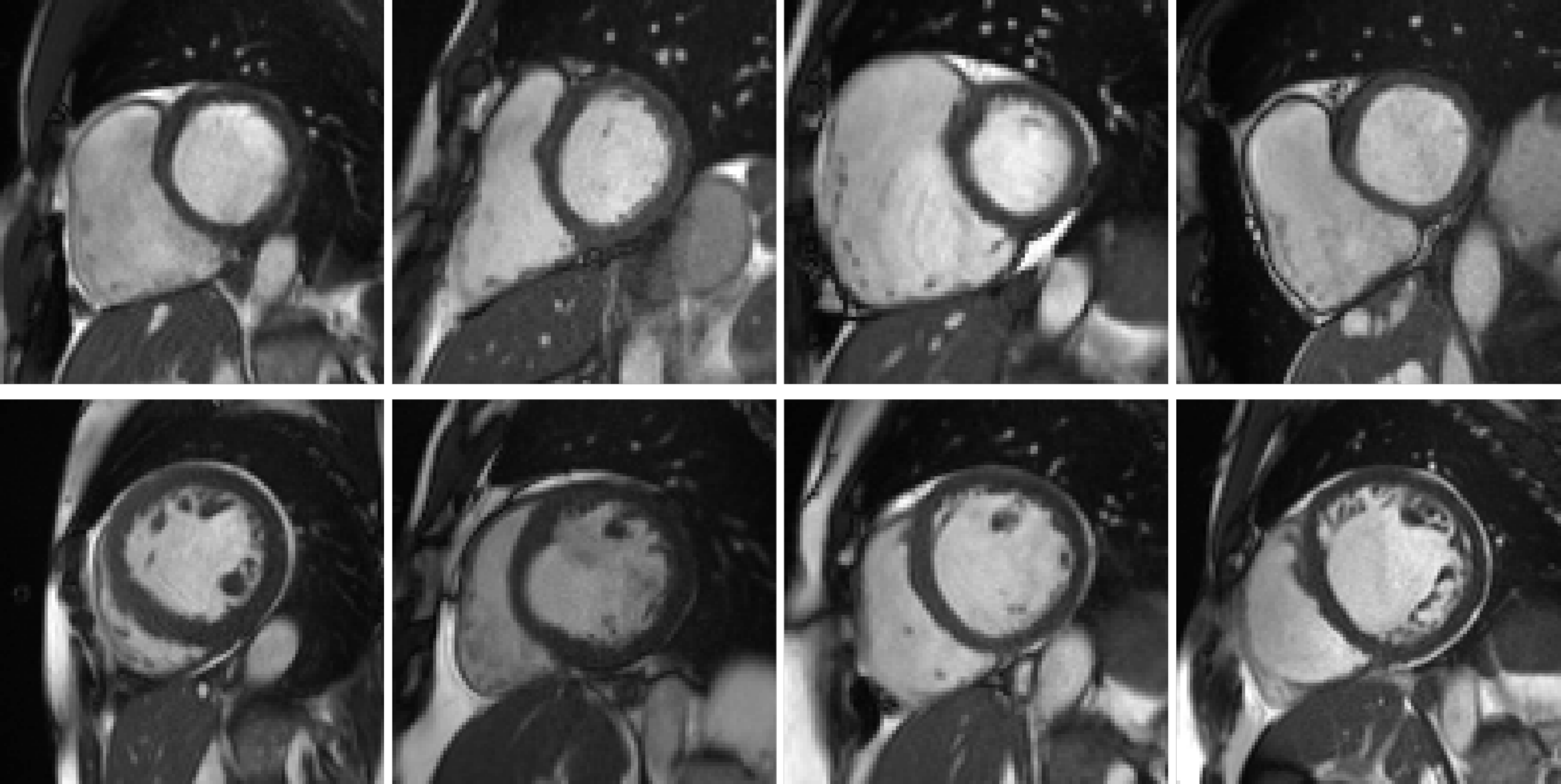}
\caption{Examples of the cases in clusters \#5 and \#8. First row: example cases in cluster \#5, of which the RVs appear to be exceptionally large. Second row: cases in cluster \#8, of which the LVs seem to be dilated.}
\end{figure*}

Among the 9 resulting clusters (termed cluster \#1 to \#9) of the selected model, two are of small sizes (clusters \#5 and \#8). We find that they actually correspond to two pathological categories according to the definition given by the ACDC challenge (RVA and DCM respectively).

Cluster \#5 has 11 cases (examples are given in Figure 3). As listed in Table 2, these cases have exceptionally large right ventricles, which are above 130 $\mathit{mL/m^2}$. In the ACDC challenge, the RVA cases are described as of RVC volumes higher than 110 $\mathit{mL/m^2}$ or RVC ejection fraction lower than 40\%. Hence according to the definition of ACDC, cluster \#5 is a group of cases belonging to RVA. 

Cluster \#8 has 4 cases (examples are given in Figure 3). As shown in Table 3, these cases have large LVC volumes at ED (above 130 $\mathit{mL/m^2}$) and low LVC ejection fractions (below 30\%). In the ACDC challenge, DCM cases are those with LVC volumes larger than 100 $\mathit{mL/m^2}$ and LVC ejection fraction lower than 40\%. So cluster \#8 is a group of DCM cases according to ACDC. In addition, we find that the ground-truth measures of LVC volume at ED and LVC ejection fraction are available for all 4 cases in UK Biobank (last two columns in Table 3). It is straightforward to see in Table 3 that the measures generated by our feature extraction method are quite close to the ground-truth.

For the other 7 clusters, which are of much larger sizes (above 70), we do not identify any clear correspondence between them and the pathological categories defined in the ACDC challenge.

\subsection{Further Analysis for Confirmation}

To further confirm the discovered correspondence between the two small clusters and the two pathological categories, as well as to verify whether the large clusters represent normal cases, in addition to manual verification of the segmentation masks and apparent flow maps to ensure the exactness of the features, we also conduct the following analysis.

\subsubsection{Interpretation of the Results of an ACDC Classification Model}

We apply a pathology classification model (\cite{Zheng:2018:2}) trained using the ACDC dataset on the cases of clusters \#5 and \#8.

Seven of the eleven cases of cluster \#5 are predicted to be RVA, which is as expected. However, the other 4 cases (2512949, 3423847, 4681487 and 5319688) are predicted to be NOR (i.e. normal). We suggest that this is partially due to the difference in the distributions of RVC ejection fraction. In ACDC, a great majority of the RVA cases are of RVC ejection fraction well below 50\%. So the trained model has learned to rely on this feature to determine RVA cases. Yet in UK Biobank, some RVA cases, including the 4 listed above, are of RVC ejection fraction above 50\%. They are not as severe cases as in ACDC. 

All four cases of cluster \#8 are predicted to be DCM by the classification model, which supports the correspondence between cluster \#8 and DCM. In addition, by manually checking the motion, we can confirm areas of hypokinesia and akinesia for these cases but also dyskinesia for one case (3378112). For case ID 2432774, we also observe discoordinate movement of the LV myocardium suggestive of bundle branch block, which is a type of electrical conduction disease commonly associated with structural heart disease and heart failure. These observations suggest that these cases might also have some relation to MINF. In fact, as pointed out in the ACDC challenge, the increase of LVC volume can be a consequence of the adaptation of LV due to MINF (also called cardiac remodeling).

\subsubsection{Reduced Dimensionality Visualization Using Principal Component Analysis}

\begin{figure*}[t]
\centering
\subfigure{
\includegraphics[width=8.1cm, height=6.3cm]{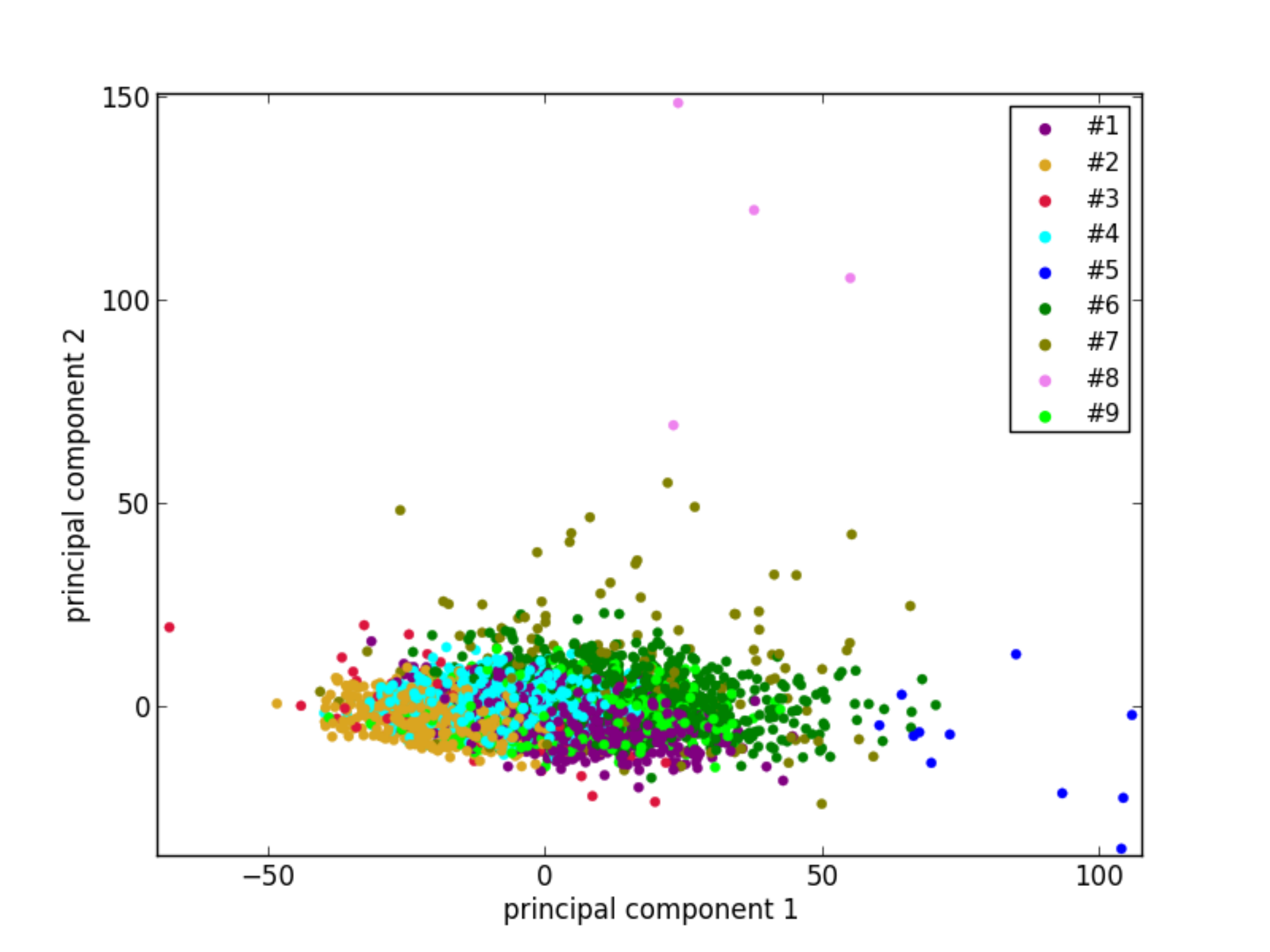}}
\subfigure{
\includegraphics[width=8.1cm, height=6.3cm]{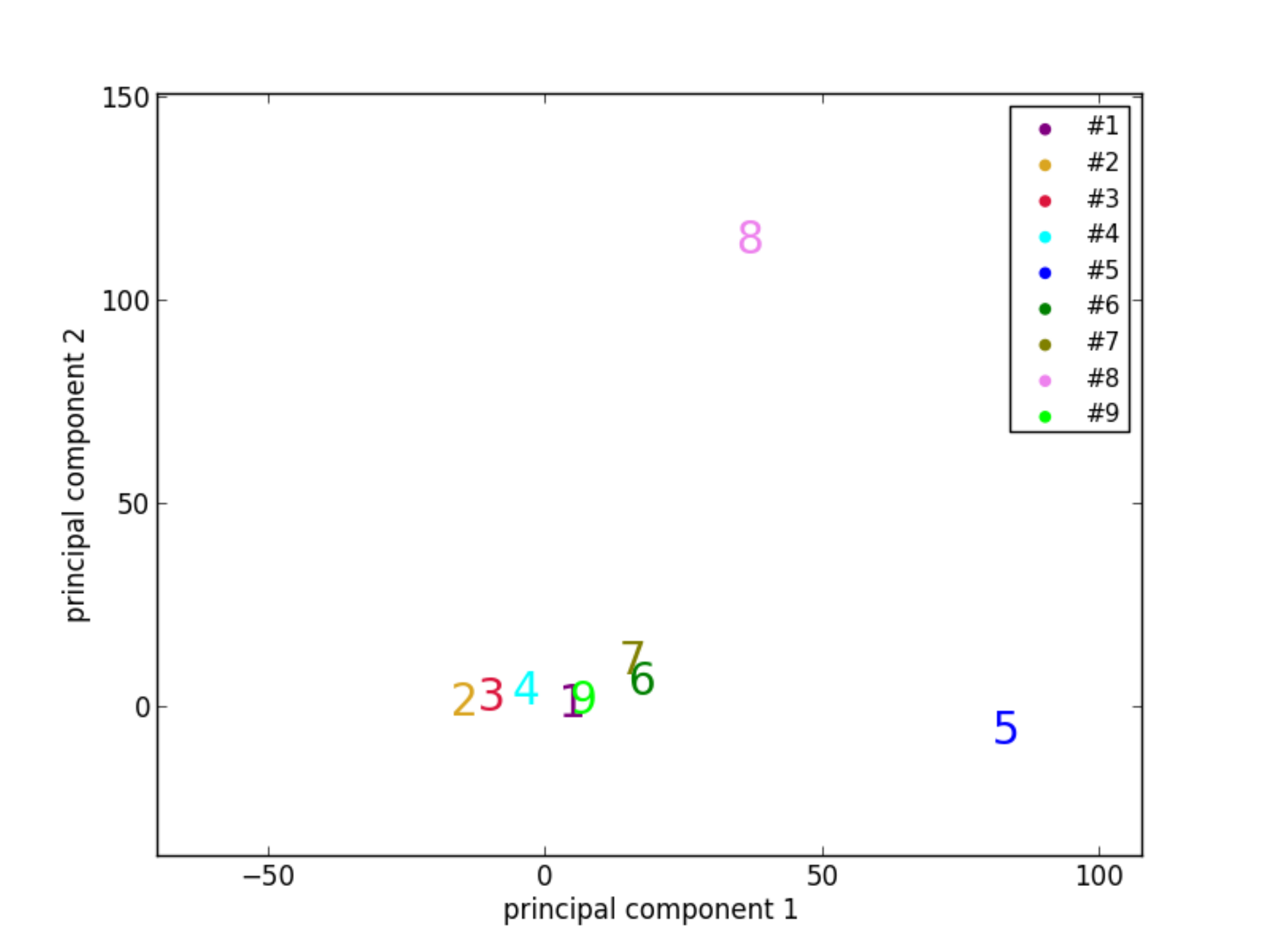}}
\caption{The results of dimensionality reduction by principal component analysis. (Left) The data points of the 3822 UK Biobank cases projected to the space of the 2 principal components. Each data point is colored according to its cluster. (Right) Projection of the centers (marked by the corresponding indexes and colors) of the 9 clusters to the same space.}
\end{figure*}

To better visualize the two isolated clusters (\#5 and \#8), we perform a principal component analysis to reduce the dimensionality of the 3822 vectors of size 8 (8 selected features of 3822 cases) of UK Biobank to 2. Furthermore, the centers of the 9 clusters are also projected to the sample space of the 2 principal components. As can be seen in Figure 4, the points corresponding to the cases of clusters \#5 and \#8, as well as the centers of the two clusters, are indeed located far away from most of the other points. This supports the suggestion that the cases in clusters \#5 and \#8, which are pathological, are quite different from most of the cases in the general population.  

\subsubsection{Visualization using t-SNE}

\begin{figure*}[]
\centering
\subfigure{
\includegraphics[width=8.1cm, height=6.3cm]{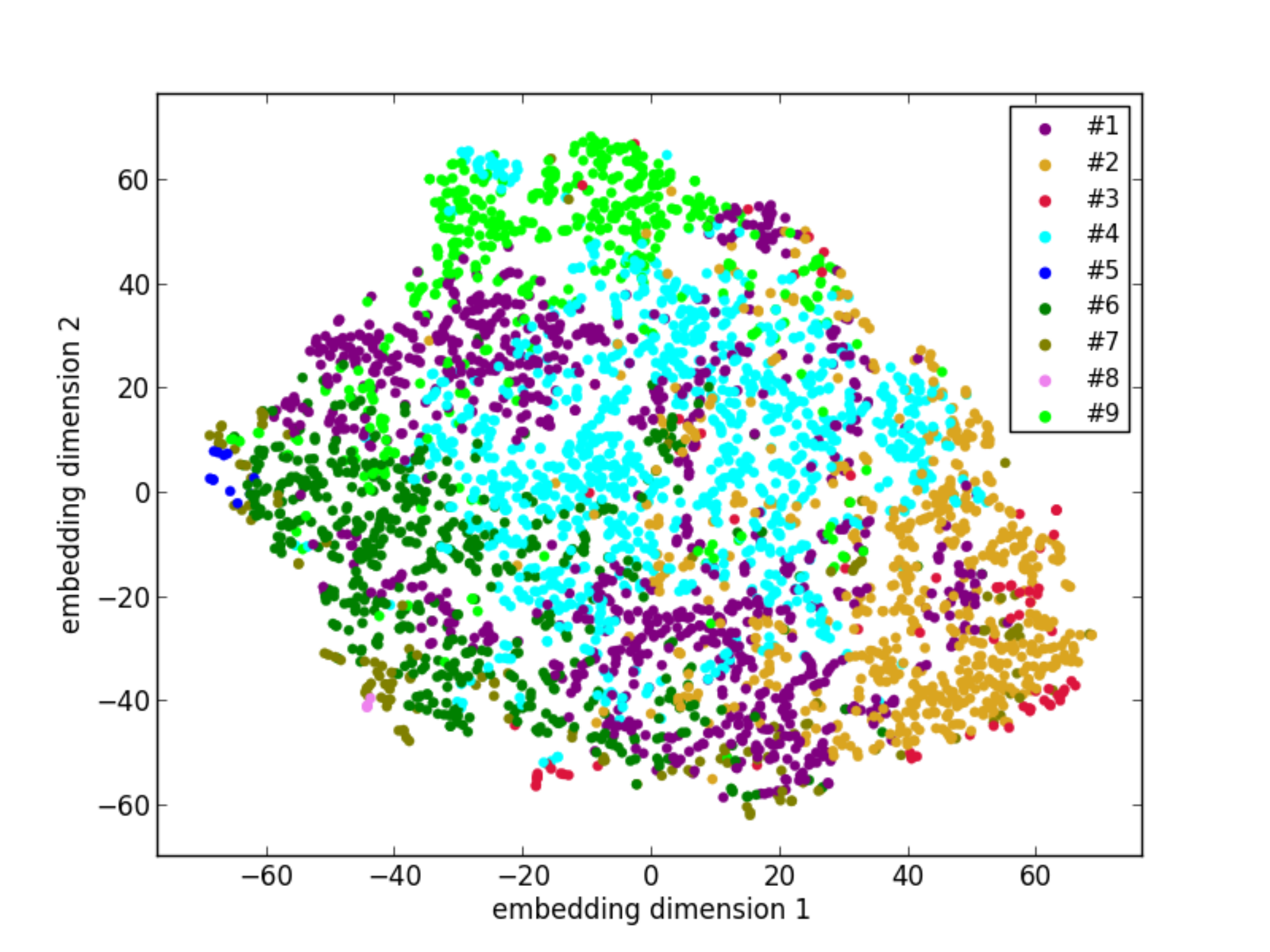}}
\subfigure{
\includegraphics[width=8.1cm, height=6.3cm]{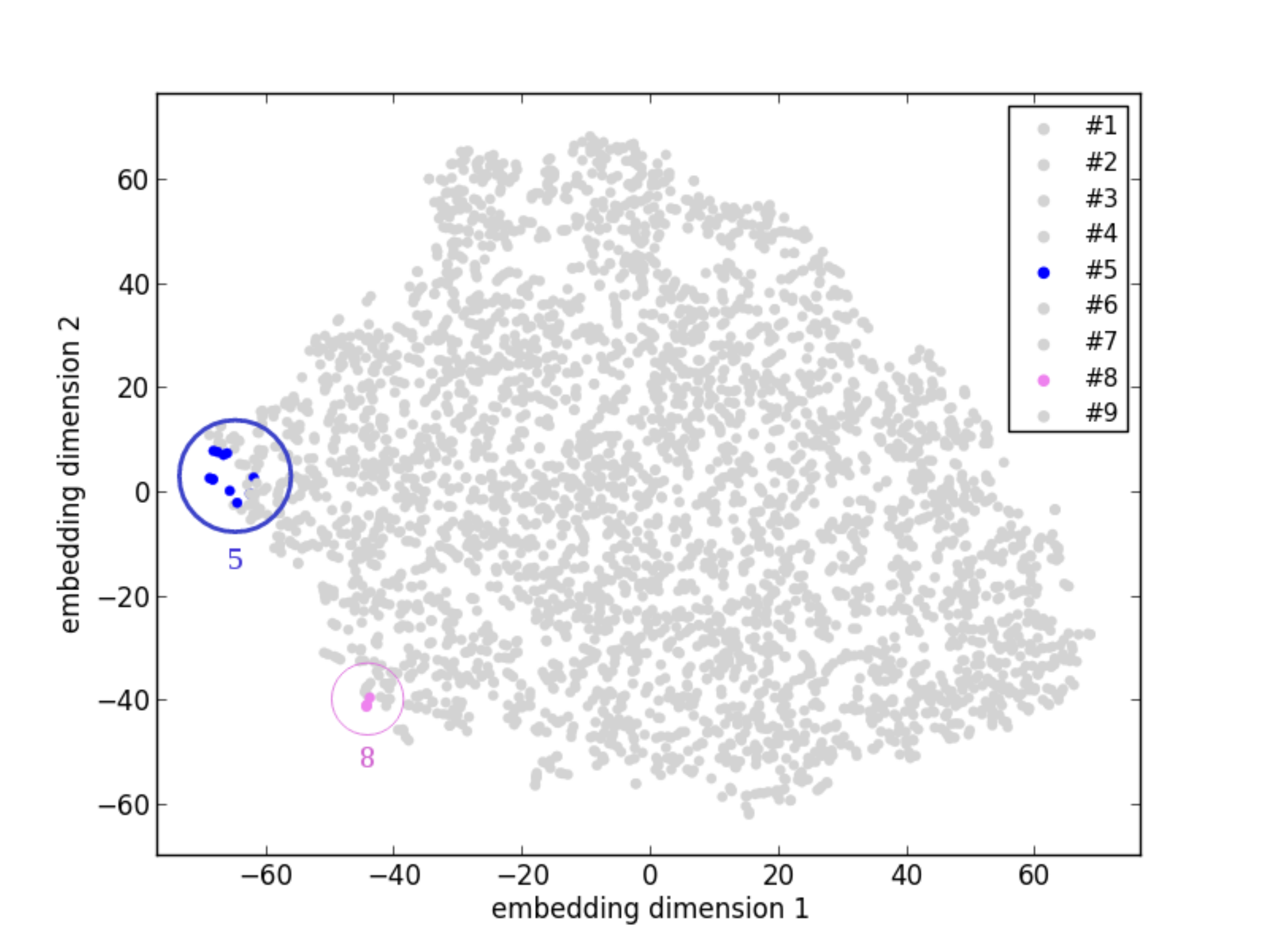}}
\caption{The results of dimensionality reduction by t-SNE. (Left) The data points of the 3822 UK Biobank cases in the space of the 2 embedding dimensions after t-SNE. Each data point is colored according to its cluster. (Right) A plot similar to the left one with only differences on coloring. Only the points of clusters \#5 and \#8 are highlighted with colors and circles.}
\end{figure*}

Similarly, another tool to visualize high-dimensional data called t-SNE (t-distributed stochastic neighbor embedding, \cite{Maaten:2008}) is applied. Its main advantage is the ability to preserve local structure. So roughly speaking, points which are close to one another in the high-dimensional space will still be close to one another after the dimensionality reduction. t-SNE is applied to the set of the 3822 vectors of the UK Biobank cases, as well as to the set of 3831 vectors which consists of the 3822 UK Biobank cases and the 9 cluster centers. Before applying t-SNE, a normalization is performed for each feature of the original data. The purpose is to make sure that each feature is on the same scale and hence has the same importance in t-SNE. As shown in Figure 5, the points of the cases and the centers of clusters \#5 and \#8 are at the edge of the ensemble of points in the embedding space. This phenomenon is again consistent with the suggestion that clusters \#5 and \#8 correspond to pathological cases which are rather different from the other cases in the general population.

\subsubsection{Examination of the Two Largest Clusters}

\begin{figure*}[]
\centering
\includegraphics[width=16.6cm, height=17.75cm]{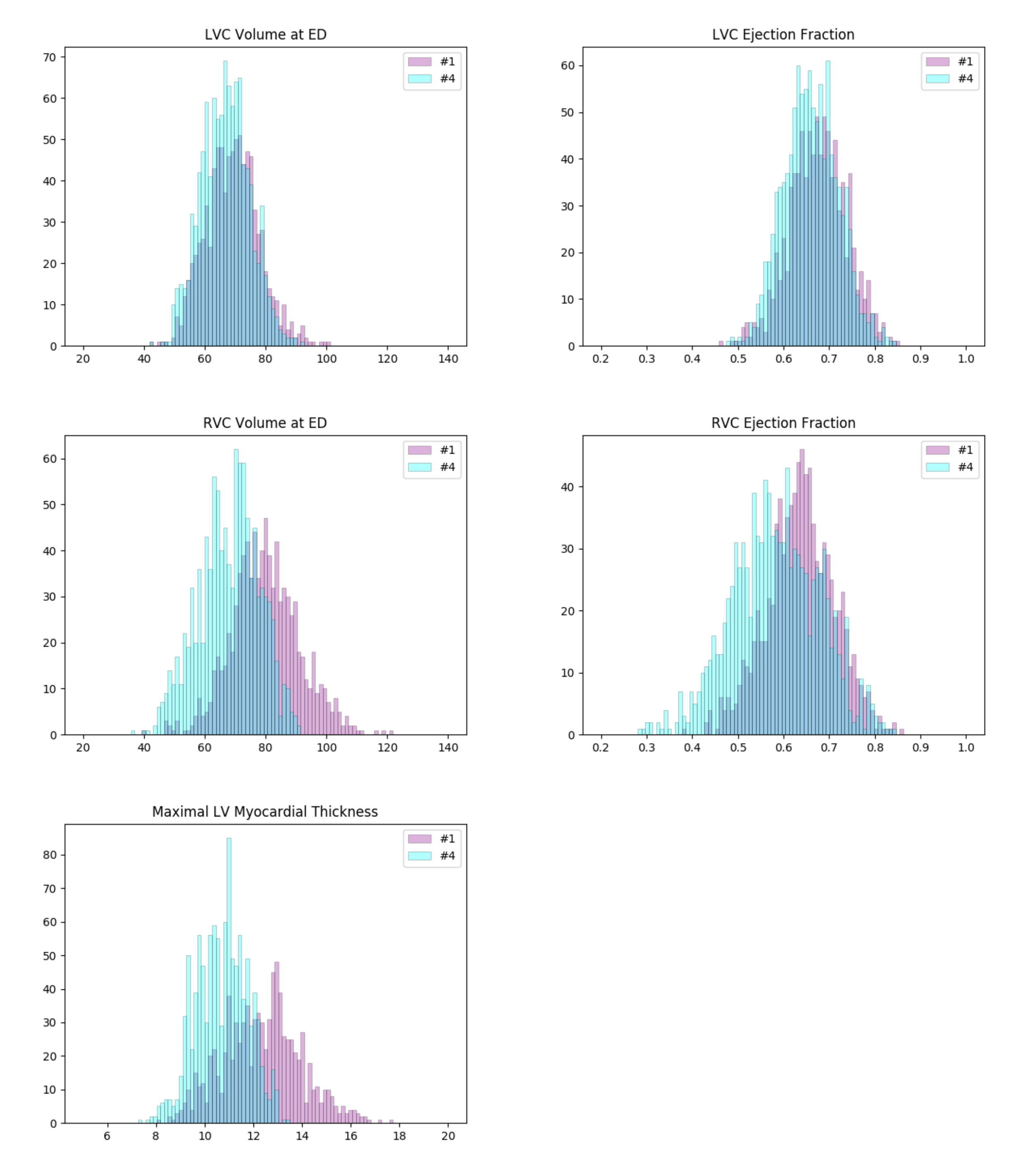}
\caption{Histograms of some important measures of the cases in clusters \#1 (pink) and \#4 (cyan). The colors of the columns are set to be partially transparent such that their overlaps appear to be of color dark blue. The distributions of \#1 and \#4 are pretty similar in terms of LVC volume and LVC ejection fraction (1st row). But they are different on RVC volume, RVC ejection fraction and maximal myocardial thickness (2nd and 3rd rows). On average, the cases of \#1 have larger RVCs with higher ejection fractions. And their myocardiums also tend to be thicker than that of the cases of \#4. For both clusters, the measures are well in normal ranges according to the definitions given by ACDC.}
\end{figure*}

As pointed out previously, while the pathological categories of clusters \#5 and \#8 are identifiable, we do not see how the other seven large clusters correspond to any cardiac pathology. In particular, the largest clusters which are of several hundreds or even more cases probably represent groups of normal cases. To verify this, we further examine the two largest clusters (\#1 and \#4, 889 and 1075 cases, respectively). 

We plot the histograms of their ventricle volumes and ejection fractions, as well as their maximal myocardial thicknesses (Figure 6). The distributions of \#1 and \#4 look pretty similar in terms of LVC volume and LVC ejection fraction. But they are different on RVC volume, RVC ejection fraction and maximal myocardial thickness. On average, the cases of \#4 have larger RVCs with higher ejection fractions. And their myocardiums also tend to be thicker than that of the cases of \#1. Furthermore, we perform the unpaired unequal variance t-test to prove that the corresponding means of the distributions of \#1 and \#4 are different. Under the null hypotheses that the corresponding distributions have the same mean, the p-values for LVC volume, LVC ejection fraction, RVC volume, RVC ejection fraction and maximal myocardial thickness are all much below 0.05 (lower than $10^{-7}$), which are small enough to reject the null hypotheses. This means that clusters \#1 and \#4 actually exhibit significant different values of the 5 features (LVC volume at ED, LVC ejection fraction, RVC volume at ED, RVC ejection fraction and maximal myocardial thickness).

For both clusters, at least a great majority of the cases satisfy:\\
\textbullet \ LVC volumes at ED less than 100 $\mathit{mL/m^2}$\\
\textbullet \ LVC ejection fraction above 40\%\\
\textbullet \ RVC volumes at ED less than 110 $\mathit{mL/m^2}$\\  
\textbullet \ RVC ejection fraction above 40\%\\
\textbullet \ Maximal myocardial thickness less than 15 mm\\
Hence according to the definitions in ACDC, these two clusters do not correspond to any of the 4 pathological categories (DCM, HCM, MINF, RVA).

\subsubsection{Examination of the Seven Large Clusters}

\begin{table*}[]
\caption{The large p-values of the unpaired unequal variance t-tests for the 21 pairs of clusters in the seven large clusters, and for the 8 extracted features, under the null hypothesis that the distributions of the feature has the same mean for both clusters. For most of the cluster pairs and features, the p-values are below 0.05}
\centering
\begin{tabular}{c c}
\hline
\noalign{\vskip 0.0in}
\multicolumn{1}{|c|}{cluster pair} & \multicolumn{1}{c|}{p-values above 0.05 (and the corresponding features)}\\  
\hline
\multicolumn{1}{|c|}{(\#1, \#4)} & \multicolumn{1}{c|}{0.07 ($V_\mathit{LVC,ES}$)}\\ 
\hline
\multicolumn{1}{|c|}{(\#1, \#6)} & \multicolumn{1}{c|}{0.56 ($\mathit{RMD}$), 0.05 ($TMD$)}\\ 
\hline
\multicolumn{1}{|c|}{(\#1, \#9)} & \multicolumn{1}{c|}{0.55 ($V_\mathit{RVC,ED}$), 0.76 ($\mathit{R_{RVCLV,ED}}$)}\\ 
\hline
\multicolumn{1}{|c|}{(\#2, \#3)} & \multicolumn{1}{c|}{0.17 ($\mathit{R_{RVCLV,ED}}$), 0.80 ($\mathit{R_{LVMLVC,ED}}$)}\\ 
\hline
\multicolumn{1}{|c|}{(\#2, \#4)} & \multicolumn{1}{c|}{0.31 ($TMD$)}\\ 
\hline
\multicolumn{1}{|c|}{(\#2, \#7)} & \multicolumn{1}{c|}{0.85 ($\mathit{R_{RVCLV,ED}}$), 0.76 ($\mathit{RMD}$)}\\ 
\hline
\multicolumn{1}{|c|}{(\#3, \#4)} & \multicolumn{1}{c|}{0.29 ($\mathit{R_{RVCLV,ED}}$)}\\ 
\hline
\multicolumn{1}{|c|}{(\#3, \#6)} & \multicolumn{1}{c|}{0.12 ($\mathit{EF_{RVC}}$)}\\ 
\hline
\multicolumn{1}{|c|}{(\#3, \#7)} & \multicolumn{1}{c|}{0.07 ($\mathit{EF_{RVC}}$), 0.28 ($\mathit{R_{RVCLV,ED}}$),}\\ 
\multicolumn{1}{|c|}{} & \multicolumn{1}{c|}{0.61 ($\mathit{MT_{LVM,ED}}$), 0.25 ($TMD$)}\\ 
\hline
\multicolumn{1}{|c|}{(\#4, \#6)} & \multicolumn{1}{c|}{0.70 ($\mathit{R_{LVMLVC,ED}}$), 0.14 ($TMD$)}\\ 
\hline
\multicolumn{1}{|c|}{(\#6, \#7)} & \multicolumn{1}{c|}{0.27 ($\mathit{EF_{RVC}}$)}\\ 
\hline
\end{tabular}
\end{table*}

To further understand the seven large clusters, we first systematically perform the unpaired unequal variance t-test. For each pair of clusters in the seven large clusters, and for each of the 8 extracted features, under the null hypothesis that the distributions of the feature has the same mean for both clusters, the p-value is computed. In this way 21$\times$8=168 p-values are obtained. In total, 149 p-values among them are below 0.05, which are small enough to reject the corresponding null hypotheses. This confirms that the clusters have different distributions on the features. Nineteen p-values among them are above 0.05, which signify a kind of similarity between pairs of clusters (Table 4). Similarly, we perform the unpaired two-sided Mann-Whitney rank tests, under the null hypotheses that the corresponding distributions of the features are the same for both clusters. And we find again that a great majority (147) of the p-values are below 0.05 such that the corresponding null hypotheses can be rejected.

\subsubsection{Measures by the Automatic Pipeline versus the Ground-Truth}

\begin{table*}[]
\caption{The means and standard deviations of the measures (in $\mathit{mL/m^2}$) by the automatic pipeline versus the ground-truth.}
\centering
\begin{tabular}{c c c}
\hline
\noalign{\vskip 0.0in}
\multicolumn{1}{|c}{} & \multicolumn{1}{|c|}{Automatic pipeline} & \multicolumn{1}{c|}{Ground-truth} \\  
\hline
\multicolumn{1}{|c}{LVC volume at ED ($\mathit{mL/m^2}$)} & \multicolumn{1}{|c|}{70.56 (13.91)} & \multicolumn{1}{c|}{75.48 (28.62)} \\ 
\hline
\multicolumn{1}{|c}{LVC volume at ES ($\mathit{mL/m^2}$)} & \multicolumn{1}{|c|}{24.06 (9.02)} & \multicolumn{1}{c|}{33.87 (22.82)} \\ 
\hline
\multicolumn{1}{|c}{LVC ejection fraction} & \multicolumn{1}{|c|}{66.41\% (7.33\%)} & \multicolumn{1}{c|}{56.04\% (6.53\%)} \\ 
\hline
\end{tabular}
\end{table*}

\begin{figure*}[t]
\centering
\includegraphics[width=16.8cm, height=7.0cm]{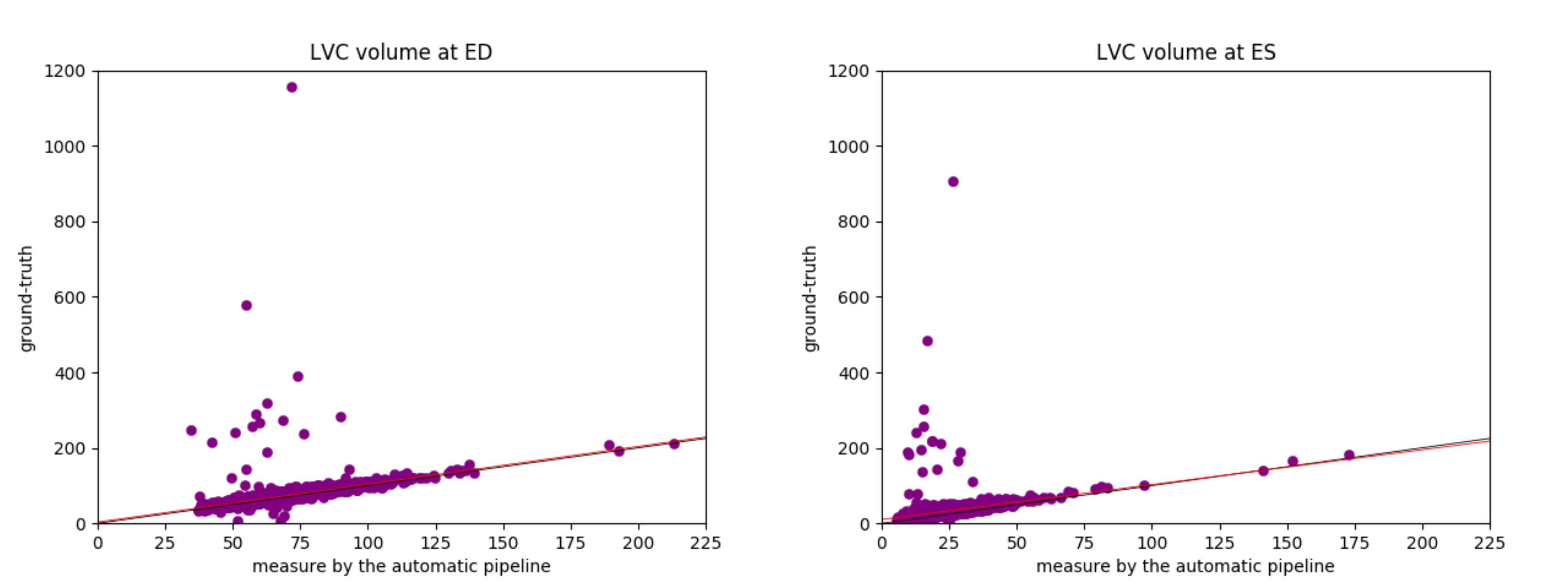}
\caption{The plots of the measures (in $\mathit{mL/m^2}$) generated by the automatic pipeline against the ground-truth for the LVC volume at ED (left) and at ES (right). We can see that the ground-truth values contain some obvious outliers, which are often of values well above the realistic range of LVC volumes. This explains the fact that the ground-truth volumes have higher means and larger standard deviations than those estimated by the automatic pipeline. The lines corresponding to the robust linear regression models (red) and the lines corresponding to \textit{ground-truth=automatic-pipeline} (black) are also plotted. The red line and the black line almost overlap with each other.}
\end{figure*}

As mentioned previously, for part of the UK Biobank cases, the ground-truth measures given by the InlineVF analysis algorithm of LVC volumes at ED and ES and LVC ejection fraction are available. In particular, among the 3822 cases used in this paper, we have access to all of the three ground-truth measures for 3212 cases. The comparison between the means and standard deviations of the measures generated by the automatic pipeline used in this paper and the ground-truth measures are shown in Table 5. It is clear that the ground-truth measures of the volumes are higher and of larger standard deviations than those estimated by the automatic pipeline. 

To better understand the cause of these differences, we plot the points of the measures in Figure 7. We can see that the ground-truth values contain some obvious outliers, which are often of values well above the realistic range of LVC volumes. This explains the fact that the ground-truth volumes have higher means and larger standard deviations than those estimated by the automatic pipeline. Moreover, proportionally, the mean of the ground-truth values of LVC volume at ED is 7.0\% (= 75.48/70.56 - 1) above that of the estimates by the automatic pipeline, while for LVC volume at ES the ground-truth is on average 40.8\% (= 33.87/24.06 -1) higher than the values obtained via the automatic pipeline. This also explains why the ground-truth of LVC ejection fraction is on average lower than that given by the automatic pipeline. The models obtained by the robust linear regression using Huber's criterion for LVC volume at ED and ES are \textit{ground-truth=1.002$\times$automatic-pipeline+3.373} and \textit{ground-truth=0.923$\times$automatic-pipeline+10.303}, respectively. The lines corresponding to the robust linear regression models (red) and the lines corresponding to \textit{ground-truth=automatic-pipeline} (black) are plotted in Figure 7. On both graphs in Figure 7, the red line and the black line almost overlap with each other. This means that our regression lines are near the lines of identity, which signifies a similarity between the measures by our method and those based on the InlineVF algorithm. By comparing the regression lines and identity lines in Fig. 4 of \cite{Suinesiaputra:2018}, we can also conclude a similarity between the measures derived from manual segmentation and those based on the InlineVF algorithm. Hence our method actually generates measures which are close to both manual and InlineVF values.

We believe that the differences between the measures by the automatic pipeline used in this paper and the ground-truth are partially due to the lack of quality control on the ground-truth. In fact, as pointed out in \cite{Suinesiaputra:2018}, the ground-truth is generated by the InlineVF algorithm, which may fail and hence make unreliable predictions on some cases. Without quality control, these failures causes the outliers in Figure 7.

\section{Conclusion and Discussion}

In this paper, we proposed a method of unsupervised cluster analysis on a large unlabeled dataset (UK Biobank) of the general population to identify pathological cases based on shape-related and motion-characteristic features extracted from cardiac cine MRI images. As far as we know, this is a topic that has rarely been studied before. In our cluster analysis, a Gaussian mixture model is applied to cluster similar cases together without supervision. As a result, among the generated clusters, we identify two that probably correspond to two cardiac pathological categories. This idea is further supported by the observations on the results of a trained classification model and of the dimensionality reduction tools including principal component analysis and t-SNE.

As more and more large and unlabeled datasets are available in the community, researchers will be able to extract interesting information by data mining. Identification of cardiac pathology is just one among other topics such as the analysis of motion patterns, the relationship between motion and shape features, etc. In the future, more research may be carried out by including more data and different types of data (\cite{Kohli:2017}), using more features, targeting other abnormalities or phenotype properties, etc. Various unsupervised learning methods (\cite{Raza:2018}) other than a Gaussian mixture model can also be applied.

\section*{Conflict of interest}
The authors declare that they have no conflicts of interest.

\section*{Acknowledgments}

The authors acknowledge the partial support from the European Research Council (MedYMA ERC-AdG-2011-291080). This research has been conducted using the UK Biobank Resource under application 2964. Funding was provided by British Heart Foundation (PG/14/89/31194). The authors also thank Stefan K. Piechnik, Stefan Neubauer, Nay Aung, Jose M. Paiva, Aaron M. Lee, Elena Lukaschuk, Mihir Sanghvi, Mohammed Y. Khanji, Filip Zemrak, Valentina Carapella and Young Jin Kim for contributing in the manual analysis of the UK Biobank cases. Steffen E. Petersen acknowledges support from the NIHR Barts Biomedical Research Centre and from the ``SmartHeart'' EPSRC programme grant (EP/P001009/1). Steffen E. Petersen provides consultancy to Circle Cardiovascular Imaging Inc., Calgary, Alberta, Canada.

\section*{References}

\bibliography{mybibfile}

\end{document}